\documentclass[runningheads]{llncs}
\usepackage[T1]{fontenc}
\usepackage{graphicx}
\usepackage {booktabs}
\usepackage{amsmath,amssymb}
\usepackage{bm}
\newcommand{\etal}{\textit{et al}.}

\newcommand{\eg}{\textit{e}.\textit{g}.}

\usepackage{graphicx}
\usepackage{booktabs}
\usepackage{multirow}
\usepackage{comment}
\usepackage{url}
\usepackage{color, colortbl}
\usepackage[pagebackref,breaklinks,colorlinks]{hyperref}

\definecolor{Gray}{gray}{0.9}
\newcolumntype{g}{>{\columncolor{Gray}}c}

\begin{document}
\title{EgoSurgery-Phase: A Dataset of Surgical Phase Recognition from Egocentric Open Surgery Videos}
\titlerunning{EgoSurgery-Phase}
\author{Ryo Fujii\inst{1}\orcidID{0000-0002-9115-8414} \and
Masashi Hatano\inst{1} \and
Hideo Saito\inst{1}\orcidID{0000-0002-2421-9862} \and
Hiroki Kajita\inst{2}}
\authorrunning{R. Fujii \etal}
\institute{Keio University, Yokohama, Kanagawa, Japan \\
\email{\{ryo.fujii0112, hatano1210, hs\}@keio.jp}\and
Keio University School of Medicine, Shinjuku, Tokyo, Japan\\
\email{\{jmrbx767\}@keio.jp} }

\maketitle              

\begin{abstract}
Surgical phase recognition has gained significant attention due to its potential to offer solutions to numerous demands of the modern operating room. However, most existing methods concentrate on minimally invasive surgery (MIS), leaving surgical phase recognition for open surgery understudied. This discrepancy is primarily attributed to the scarcity of publicly available open surgery video datasets for surgical phase recognition. To address this issue, we introduce a new egocentric open surgery video dataset for phase recognition, named Egosurgery-Phase. This dataset comprises 15 hours of real open surgery videos spanning 9 distinct surgical phases all captured using an egocentric camera attached to the surgeon's head. In addition to video, the Egosurgery-Phase offers eye gaze. As far as we know, it is the first real open surgery video dataset for surgical phase recognition publicly available. Furthermore, inspired by the notable success of masked autoencoders (MAEs) in video understanding tasks (\eg, action recognition), we propose a gaze-guided masked autoencoder (GGMAE). Considering the regions where surgeons’ gaze focuses are often critical for surgical phase recognition (\eg, surgical field), in our GGMAE, the gaze information acts as an empirical semantic richness prior to guiding the masking process, promoting better attention to semantically rich spatial regions. GGMAE significantly improves the previous state-of-the-art recognition method ($6.4\%$ in Jaccard) and the masked autoencoder-based method ($3.1\%$ in Jaccard) on Egosurgery-Phase. The dataset is released at \href{https://github.com/Fujiry0/EgoSurgery}{project page}.


\end{abstract}

\section{Introduction}
Automated analysis of surgical videos is indispensable for various purposes, including providing real-time assistance to surgeons, supporting education, and evaluating medical treatments. Surgical phase recognition, the recognition of the transitions of high-level stages of surgery, is a fundamental component in advancing these objectives. 
Surgical phase recognition has gained considerable attention with numerous approaches~\cite{Tobias2020MICCAI,Gao2021MICCAI,Jin2018TMI,Jin2021TMI,Twinanda2016arXiv,TwinandaTMI2017,Yi2023ACCV}. While surgical phase recognition is important across all surgical methods, the predominant focus of research endeavors has been on minimally invasive surgery (MIS), leaving open surgery phase recognition comparatively underexplored. This discrepancy primarily stems from the scarcity of publicly available large-scale open surgery datasets for phase recognition. In the surgical phase recognition for MIS, several large-scale datasets~\cite{TwinandaTMI2017,WangMICCAI2022} have been released, driving advancements in learning-based algorithms. Conversely, the absence of comparable large-scale datasets for open surgery phase recognition has significantly impeded progress in achieving accurate surgical phase recognition within the open surgery domain.

To tackle this issue, we introduce Egosurgery-Phase, the first large-scale egocentric open surgery video dataset for phase recognition. 20 videos of procedures of 10 distinct surgical types with a total duration of 15 hours conducted by 8 surgeons are collected and annotated into 9 phases. The videos have been meticulously pre-processed for de-identification. EgoSurgery-Phase offers a rich collection of video content capturing diverse interactions among individuals (\eg, surgeons, assistant surgeons, anesthesiologists, perfusionists, and nurses), varied operative settings, and various lighting conditions. Moreover, in addition to video, EgoSurgery-Phase provides eye gaze data.

\begin{figure*}[tb]
 \centering
 \resizebox{0.9\textwidth}{!}{
\begin{tabular}{c}
\begin{tabular}{cccccc}
\begin{minipage}{0.20\hsize}
    \begin{center}
        \includegraphics[clip, width=\hsize]{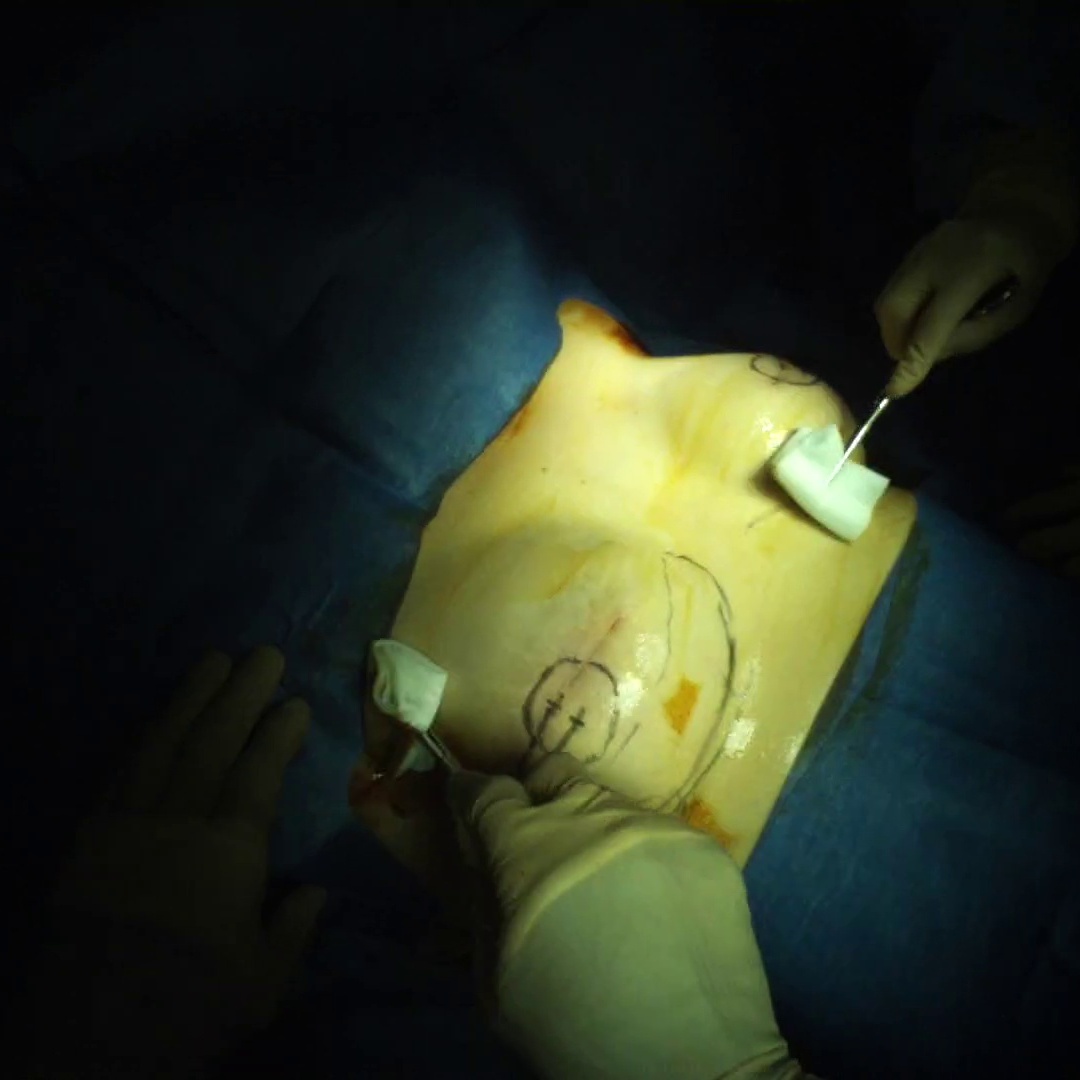}
        {\footnotesize (P1) Disinfection}
    \end{center}
\end{minipage}
&
\begin{minipage}{0.20\hsize}
    \begin{center}
        \includegraphics[clip, width=\hsize]{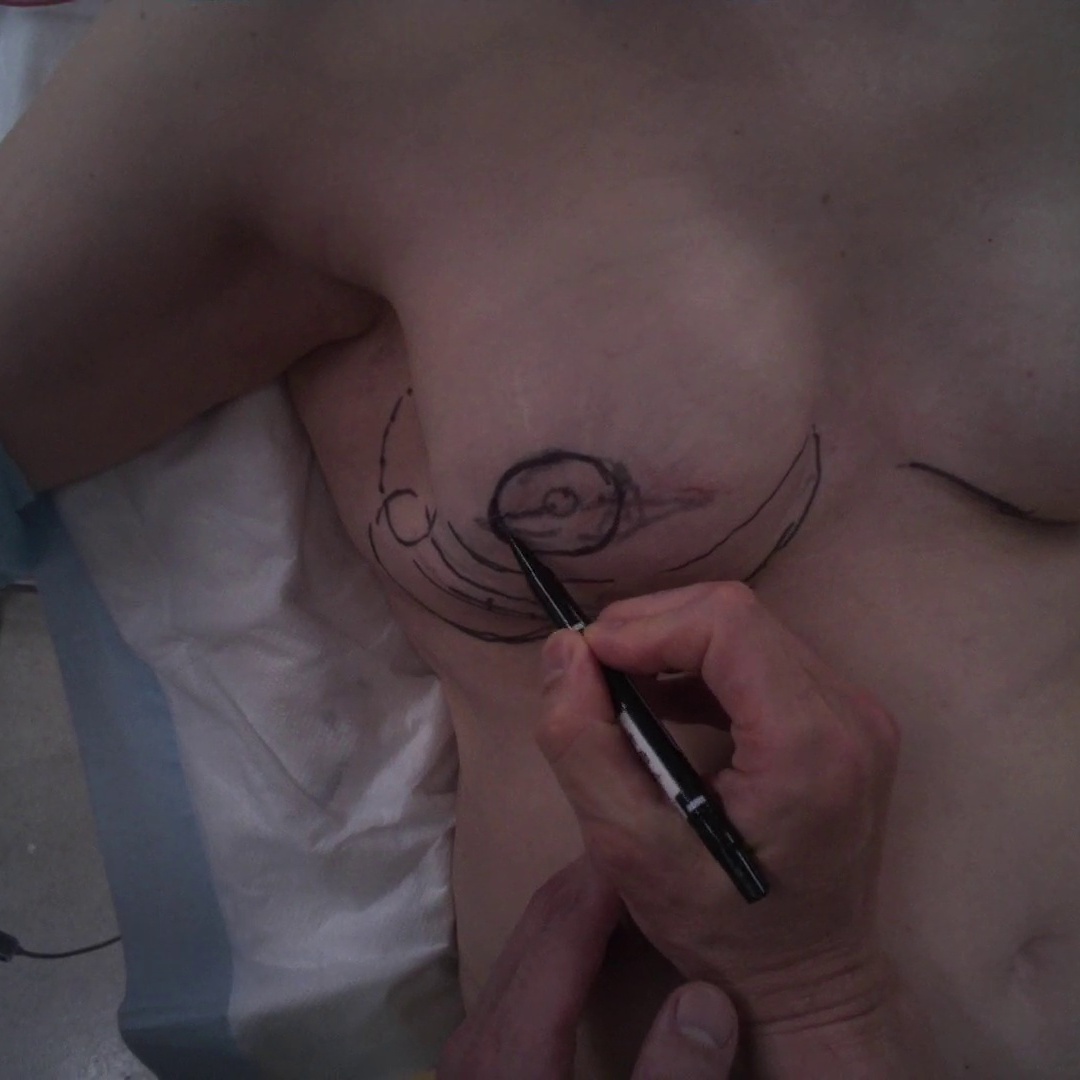}
        {\footnotesize (P2) Design}
    \end{center}
\end{minipage}
&
\begin{minipage}{0.20\hsize}
    \begin{center}
        \includegraphics[clip, width=\hsize]{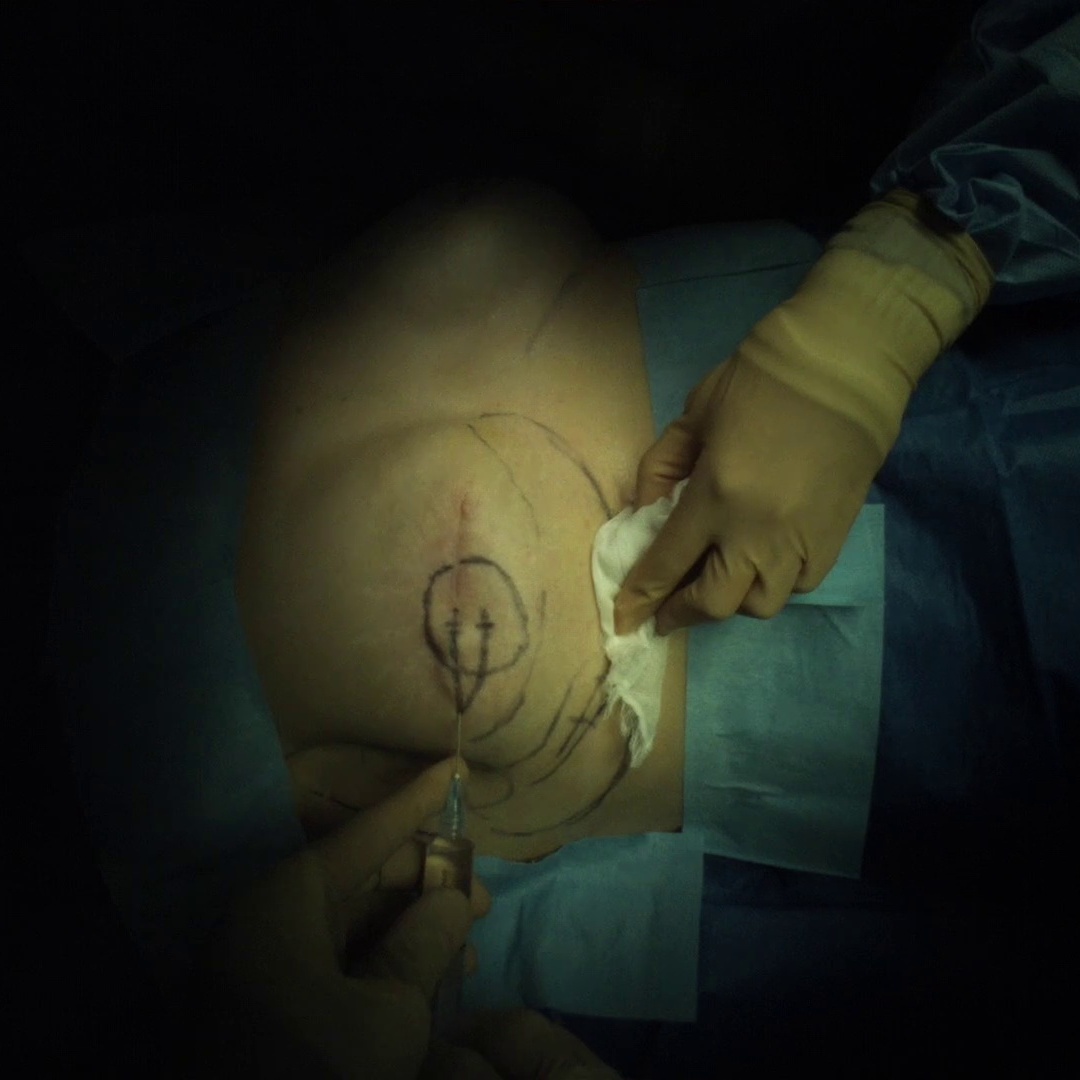}
        {\footnotesize (P3) Anesthesia}
    \end{center}
\end{minipage}
&
\begin{minipage}{0.20\hsize}
    \begin{center}
        \includegraphics[clip, width=\hsize]{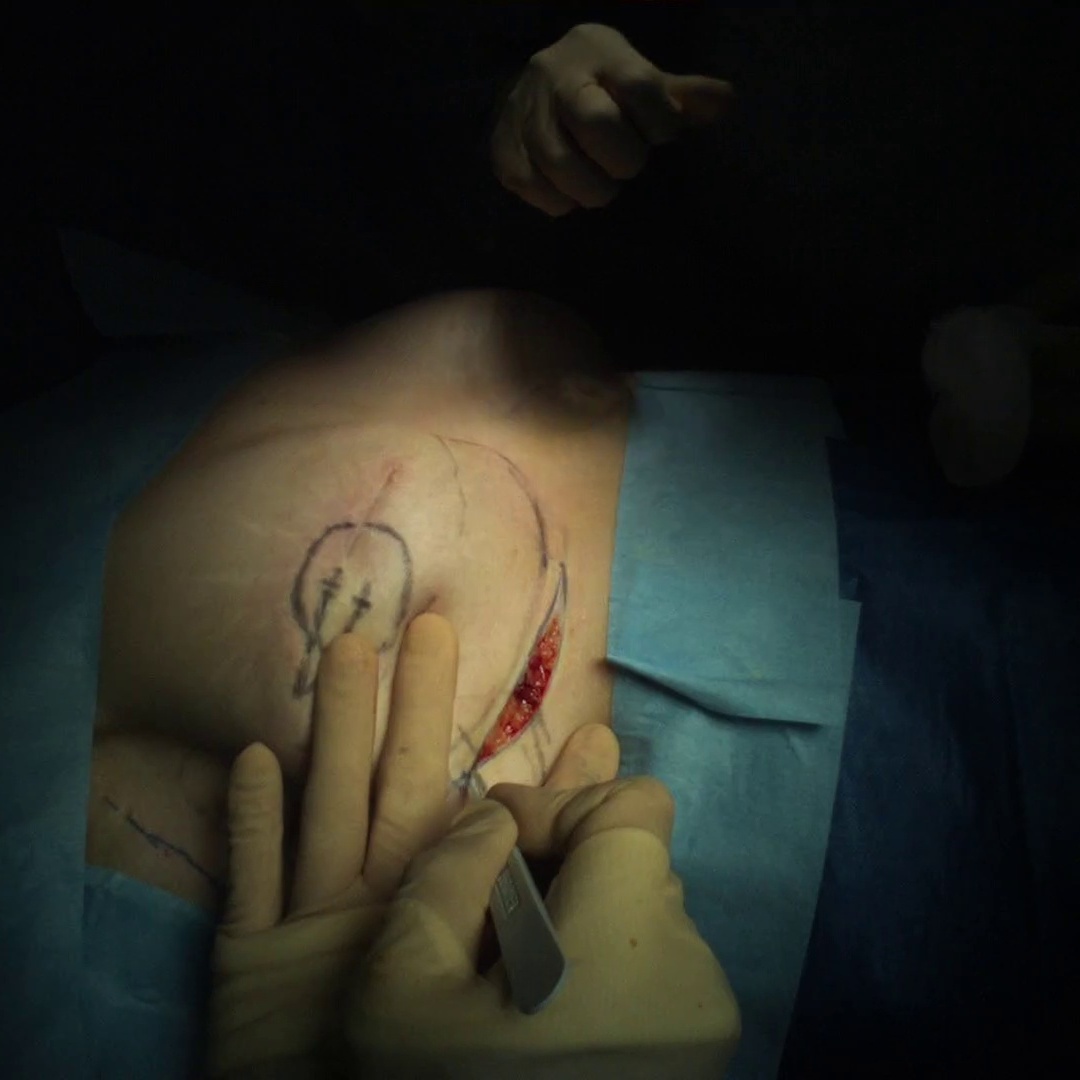}
        {\footnotesize (P4) Incision}
    \end{center}
\end{minipage}
&
\begin{minipage}{0.20\hsize}
    \begin{center}
        \includegraphics[clip, width=\hsize]{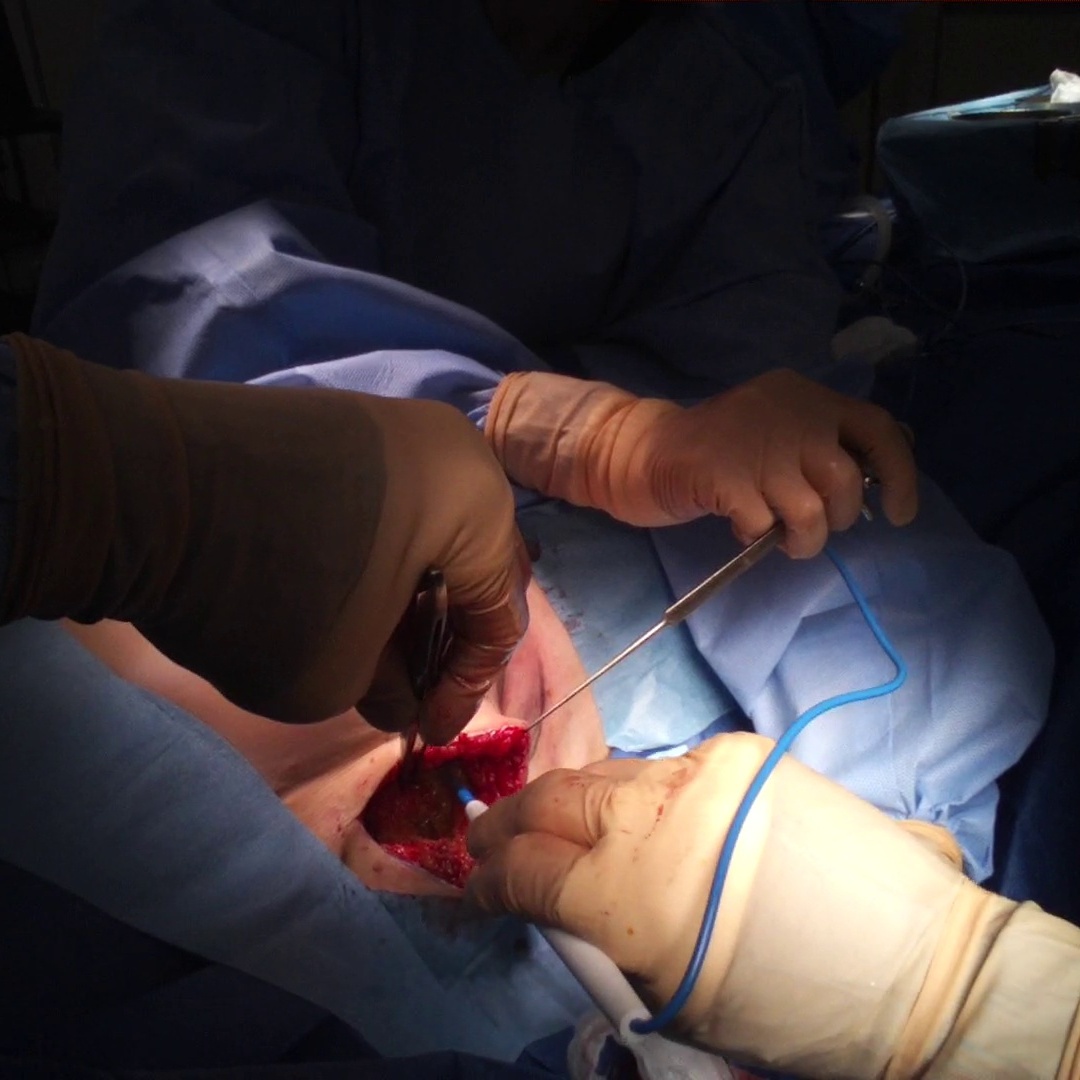}
        {\footnotesize (P5) Dissection}
    \end{center}
\end{minipage}
\end{tabular}
\vspace{1em}
\\
\begin{tabular}{ccccc}
\begin{minipage}{0.20\hsize}
    \begin{center}
        \includegraphics[clip, width=\hsize]{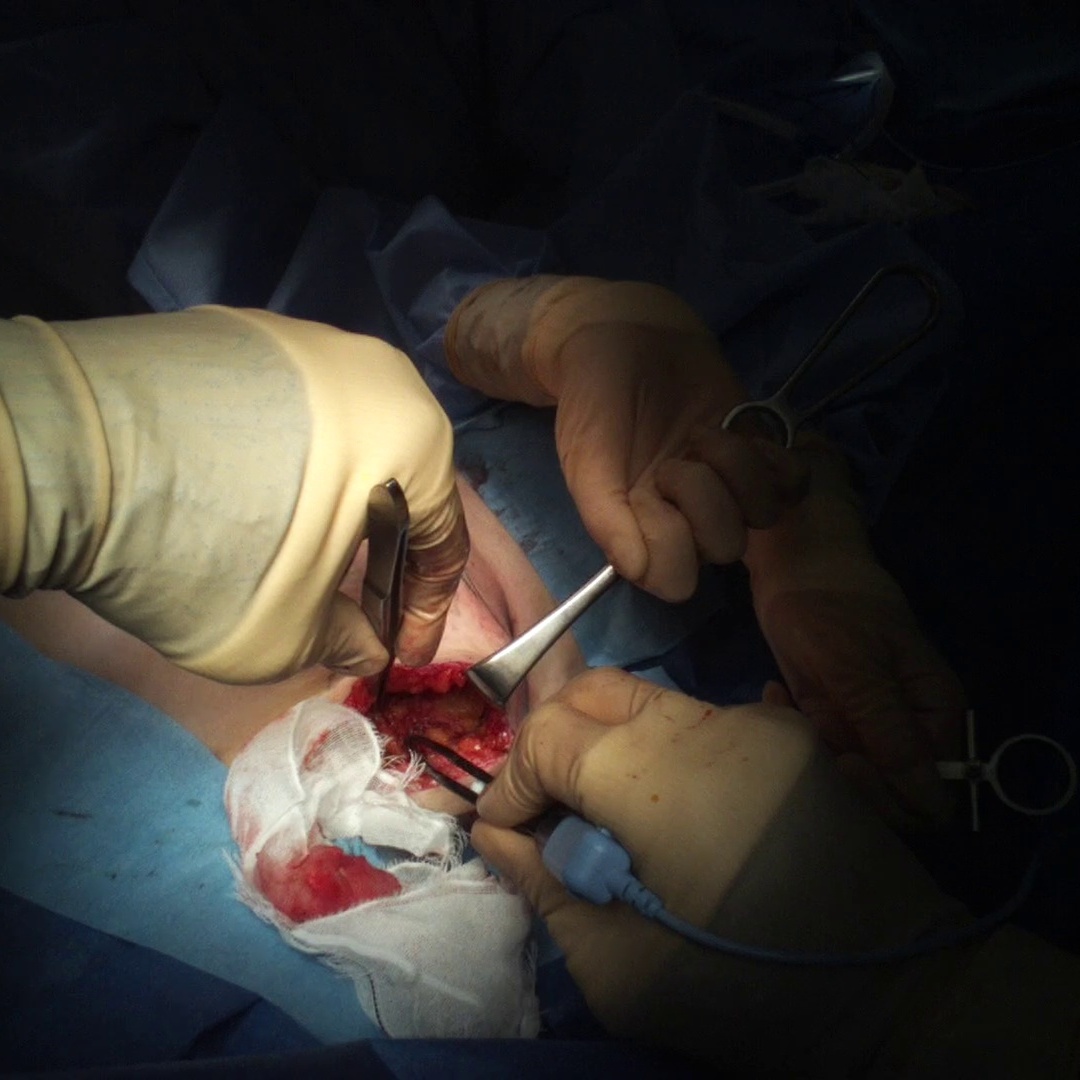}
        {\footnotesize (P6) Hemostasis}
    \end{center}
\end{minipage}
&
\begin{minipage}{0.20\hsize}
    \begin{center}
        \includegraphics[clip, width=\hsize]{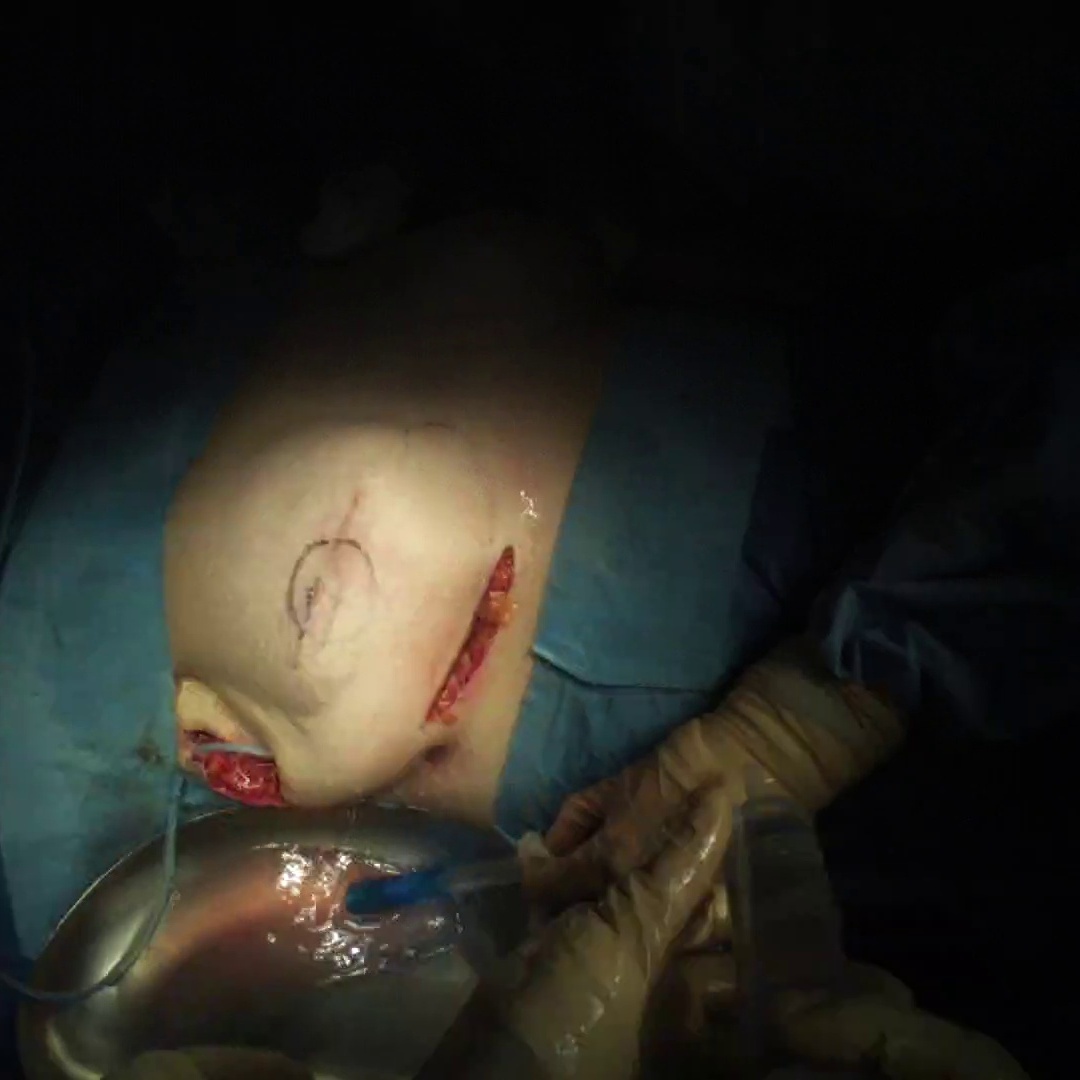}
        {\footnotesize (P7) Irrigation}
    \end{center}
\end{minipage}
&
\begin{minipage}{0.20\hsize}
    \begin{center}
        \includegraphics[clip, width=\hsize]{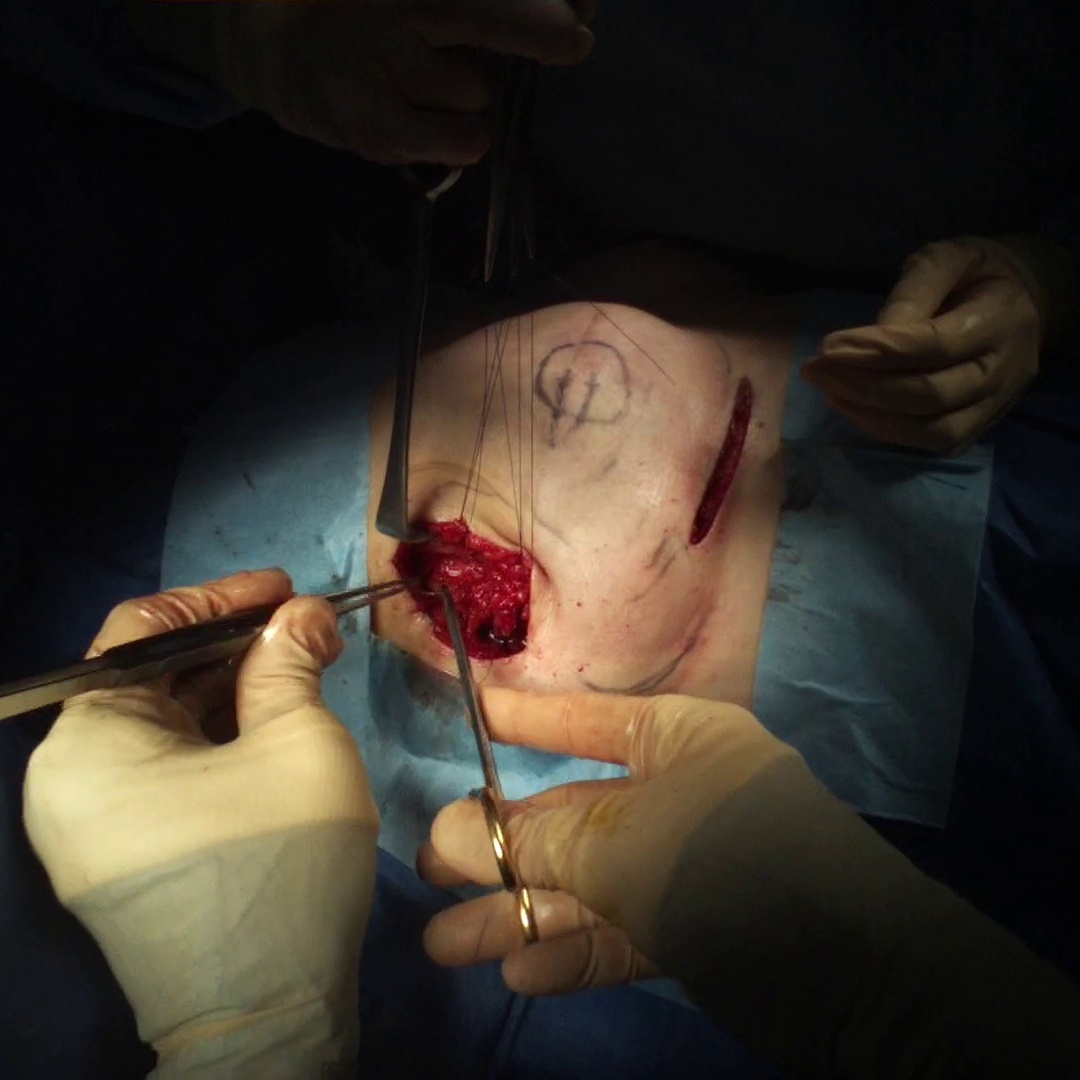}
        {\footnotesize (P8) Closure}
    \end{center}
\end{minipage}
&
\begin{minipage}{0.20\hsize}
    \begin{center}
        \includegraphics[clip, width=\hsize]{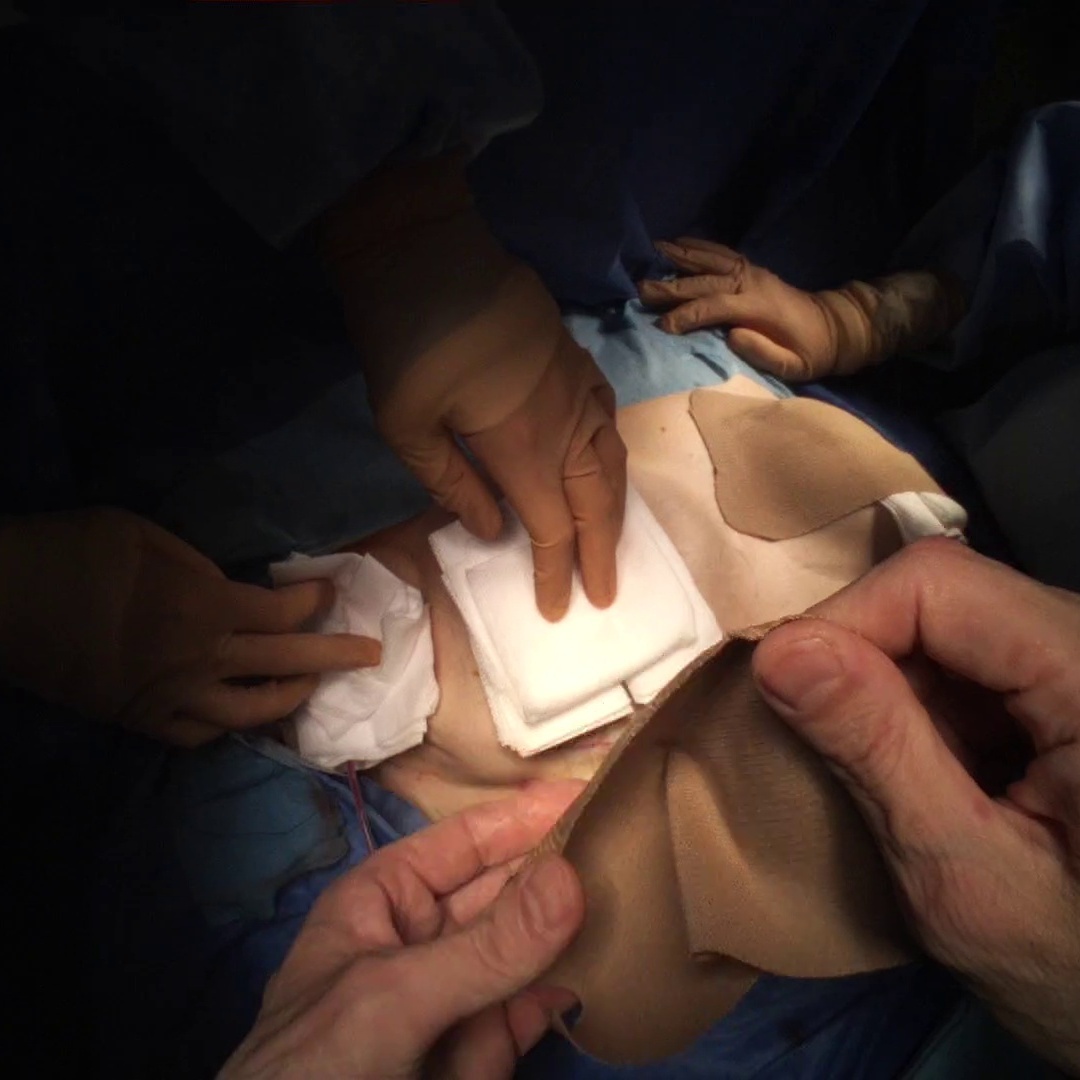}
        {\footnotesize (P9) Dressing}
    \end{center}
\end{minipage}
\end{tabular}
\end{tabular}}
\caption{Illustration of 9 surgical phases (P1-P9) annotated in the EgoSurgery-Phase dataset. Typically, the phases are executed sequentially from P1 to P9. }
 \label{fig:phaseeamples}
\end{figure*}

\begin{figure*}[tb] 
\centering
 \resizebox{0.7\textwidth}{!}{
\begin{tabular}{cccc}
RGB Image & Gaze Heatmap & Rnadom Mask  & Gaze-guided Mask \\
\begin{minipage}{0.2\hsize}
    \begin{center}
       \includegraphics[clip, width=\hsize]{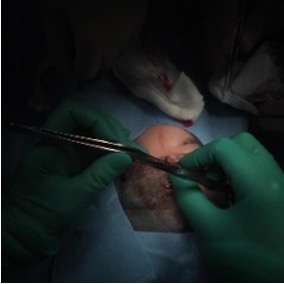}
    \end{center}
\end{minipage} & \begin{minipage}{0.2\hsize}
    \begin{center}
       \includegraphics[clip, width=\hsize]
       {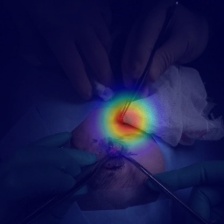}
    \end{center}
\end{minipage}  & \begin{minipage}{0.2\hsize}
    \begin{center}
       \includegraphics[clip, width=\hsize]
       {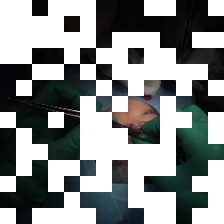}
    \end{center}
\end{minipage}  & \begin{minipage}{0.2\hsize}
    \begin{center}
       \includegraphics[clip, width=\hsize]{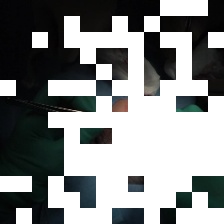}
    \end{center}
\end{minipage}  
\end{tabular}}
\caption{Example of RGB image and gaze heatmap from EgoSurgery-Phase, along with their corresponding random mask and gaze-guided mask. The gaze heatmap is depicted as a heatmap overlaid onto the RGB image for visualization purposes.}
 \label{fig:mask-examples}
\end{figure*}

Furthermore, inspired by the remarkable performance of Masked Autoencoders (MAEs)~\cite{He2022CVPR}, which learns meaningful representations by reconstructing the masked tokens, in video understanding tasks (\eg, action recognition), we propose a gaze-guided masked autoencoder (GGMAE). In MAEs, for the selection of masked tokens, a random masking strategy has been often utilized and shown to work well compared to its counterparts in some cases~\cite{He2022CVPR,Tong2022NeurIPS,Mao2023ICCV}. However, open surgery videos often contained non-informative regions (For instance, in most sample frames from EgoSurgery-Phase illustrated in Fig.~\ref{fig:phaseeamples}, we observe that the intense light from the surgical lamp causes the black clipping to outside the surgical field, making most of the tokens outside surgery field non-informative). Therefore, assuming all tokens have equal information and a uniform probability distribution for masked token selection is suboptimal. With the random masking strategy, masked tokens may be sampled from low-information regions rather than high-information ones, and training to reconstruct these tokens through MAEs is not effective~\cite{Mao2023ICCV,Sun2023CVPR}. To address this issue, we propose a gaze-guided masking approach.
Given that regions, where surgeons' gaze focuses, are often critical for surgical phase recognition (\eg, the surgical field), our GGMAE leverages gaze information as an empirical semantic richness prior to guiding the masking process, as shown in Fig. \ref{fig:mask-examples}. It converts input gaze heatmaps into a probability distribution and employs reparameterization techniques for efficient probability-guided masked token sampling. Consequently, tokens that surgeons focus on are masked with higher probability, enabling enhanced attention to semantically rich spatial regions. 

Our main contributions are summarized as follows: 1) we constructed the first publicity available large-scale real egocentric open surgery dataset, EgoSurgery-Phase, for phase recognition, 2) we propose a gaze-guided masked autoencoder, GGMAE, which incorporates gaze as an empirical semantic richness prior for masking, and 3) experimental results show that our GGMAE yields significant improvement over existing phase recognition and masked autoencoder-based methods, achieving the state-of-the-art performance on EgoSurgery-Phase.

\section{Dataset Design}
\subsection{Dataset collection}
Following the dataset collection protocol proposed in prior research~\cite{FUJII2022AS}, which focused on constructing datasets for surgical tool detection in open surgery videos, we gathered 20 open surgery videos utilizing Tobii cameras attached to the surgeon's head. The recording of patient videos received ethical approval from the Keio University School of Medicine Ethics Committee, and written informed consent was obtained from all patients or their guardians. Our dataset encompasses 10 distinct types of surgeries, performed by 8 different surgeons.

The 20 videos were recorded at a frame rate of 25 fps and a resolution of $1920 \times 1080$ pixels. Video durations vary between 28 and 234 minutes, reflecting the diversity in type and complexities of surgery. In total, 28 hours of surgical footage were captured. Unlike videos of minimally invasive surgery (MIS), open surgery videos are more likely to contain personally identifiable information (PII) such as the faces of patients, assistant surgeons, and nurses. To address privacy concerns, we subsampled the videos to 0.5 fps and anonymized the patient's face through blurring. In addition, we exclude frames containing other PII. After these pre-processing steps, the average duration of the videos becomes 46 minutes, resulting in a total duration of 15 hours, thereby yielding a large-scale dataset of high quality. In addition to video, EgoSurgery-Phase provides eye gaze.

\begin{figure}[tb]
\centering
 \resizebox{0.5\textwidth}{!}{

\begin{minipage}{\hsize}
    \begin{center}
       \includegraphics[clip, width=\hsize]{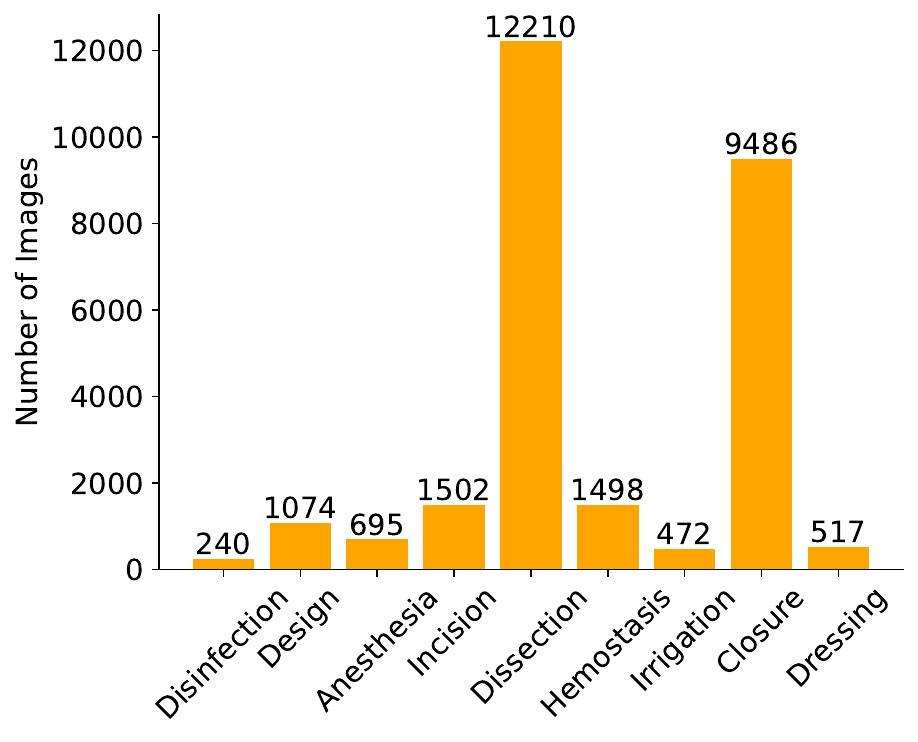}
    \end{center}
\end{minipage}

}
\caption{The phase distribution of frames.}
\label{fig:statistics}
\end{figure}

\subsection{Dataset annotation, statistics and data split}
Expert surgeons perform the annotations based on their clinical experience and domain knowledge. The 20 pre-processed videos of open surgery are manually annotated into 9 phases: Disinfection, Design, Anesthesia, Incision, Dissection, Hemostasis, Irrigation, Closure, and Dressing. Samples are shown in Fig.~\ref{fig:phaseeamples}. In total, $27,694$ frames are manually annotated. The sample distribution is shown in Fig.\ref{fig:statistics}. It reveals a notable class imbalance. We use $14$ videos for the training set, $2$ videos for the validation set, and $5$ videos for the test set.

\section{Approach}
\subsection{Overview}
\begin{figure*}[tb] 
\centering
\begin{minipage}{\hsize}
    \begin{center}
       \includegraphics[clip, width=\hsize]{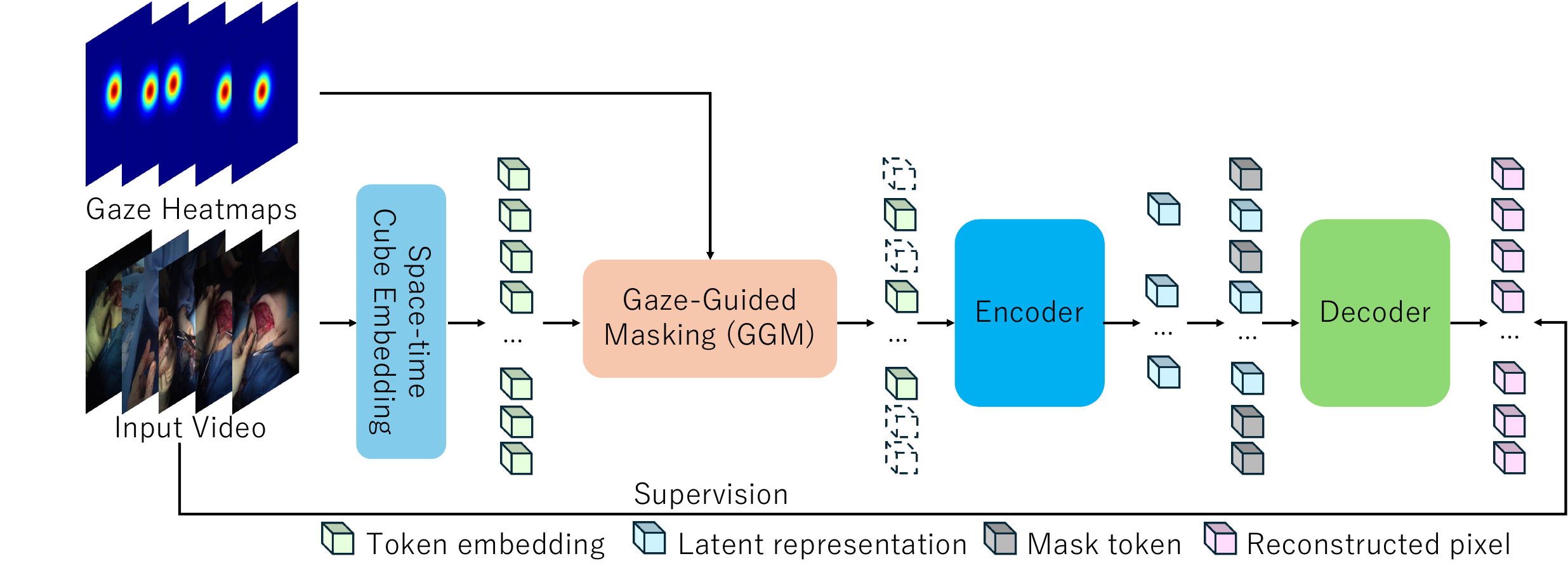}
    \end{center}
\end{minipage}
\caption{Overview of the proposed GGMAE: GGME performs the task of masking tokens and reconstructing these masked tokens with Transformer encoder-decoder architecture. Considering that open surgery videos often contain non-informative regions, we introduce the Gaze-Guided Masking (GGM) module, which selects tokens to be masked based on gaze information.}
\label{fig:overview}
\end{figure*}

Fig. \ref{fig:overview} presents an overview of the proposed GGMAE. GGMAE takes as input video $V\in\mathbb{R}^{T \times C \times H \times W}$ and gaze heatmaps $G\in\mathbb{R}^{T \times H \times W}$. Here, $C$ represents the input (RGB) channels, and $H \times W$ denotes the spatial resolution of each frame. The space-time cube embedding~\cite{Tong2022NeurIPS} is used to transform the input video into a set of token embeddings $X \in \mathbb{R}^{F \times N_sN_r}$, where $F$ is the channel dimension of the tokens, and $N_s=HW/H_cW_c$ and $N_r=T/T_c$ are the numbers of tokens along the spatial and temporal dimensions, respectively. $T_c$, $H_c$, and $W_c$ represent the size of each token along the temporal, height, and width dimensions, respectively.

We apply the proposed Gaze-Guided Masking (GGM) strategy to select tokens for masking with a masking ratio $\rho$, leveraging the gaze information. The remaining tokens, along with the space-time position embeddings, are fed into the Transformer encoder and decoder~\cite{Vaswani2017NIPS} to reconstruct the masked maps.

\subsection{Gaze-guided mask Masking}
Open surgery videos often contain non-informative regions, and training a model to reconstruct these tokens using MAE does not improve model performance~\cite{Mao2023ICCV,Sun2023CVPR}. Therefore, inspired by representation learning approaches that leverage MAEs with non-uniform masking tailored to token informativeness across diverse domain data inputs~\cite{li2022NeurIPS,Li2021NeurIPS,Mao2023ICCV,Min2023IV,Sun2023CVPR}, we integrate gaze information as an empirical semantic richness prior to guide the masking of embedding features. Specifically, we propose non-uniform token sampling based on the accumulated gaze heatmap value of each token.

First, we compute the accumulated gaze heatmap value $d_i$ for each token ${\bm x}i \in X$ by summing the heatmap values across the pixels belonging to the token as follows:
\begin{equation}
    d_i = \sum_{j \in \Omega_i} G_i,
\end{equation}
where $\Omega_i$ denotes the set of pixels in the gaze heatmap corresponding to the $i$-th token. We then calculate the masking probability vector ${\bm \pi}^t \in \mathbb{R}^{N_s}$ for each token's time index using the softmax function as follows:
\begin{equation}
    {\bm \pi}^t = {\rm Softmax}({\bm d}^t/\tau),
\label{eq:prob}
\end{equation}
where ${\bm d}^t \in \mathbb{R}^{N_s}$ represents a vector of accumulated gaze heatmap for each time index $t$, and $\tau$ is a hyper-parameter controlling the sharpness of the softmax function. Finally, the indices of the masked tokens are determined by sampling from a Multinomial distribution with probabilities ${\bm \pi}^t$, for $\lfloor \rho N_s \rfloor$ trials without replacement for each time index $t$.

\subsection{Loss function}

The loss function is the mean squared error (MSE) loss between the input pixel values and the reconstructed pixel values:
\begin{equation}
\mathcal{L} = \frac{1}{|\Omega|}\sum_{p \in \Omega} |I(p)-\hat{I}(p)|^2,
\end{equation}
where $p$ is the masked token index, $\Omega$ is the set of masked tokens, $I$ represents the input ground truth frames, and $\hat{I}$ stands for the reconstructed frames.

\section{Experiments}
\subsection{Implementation Details}
\noindent \textbf{Network Architecture.}
We employ the VideoMAE with the ViT-Small~\cite{AlexanderICLR2021} backbone. Following VidoeMAE~\cite{Tong2022NeurIPS}, we use the same input patch size of $2 \times 16 \times 16$ ($T_c \times H_c \times W_c$) for all models. We utilize 10-frame clips ($T$) as input, maintaining a fixed spatial resolution of $224 \times 224$ ($H \times W$) across all experiments. To generate the ground-truth gaze
heatmaps, we place a Gaussian centered on the ground truth gaze point.

\noindent \textbf{Pre-training details.} During pre-training, the masking ratio of the input token is set to $90\%$. We adopt the AdamW~\cite{Ilya2018ICLR} optimizer with a weight decay of $1e^{-4}$ and betas of (0.9, 0.95). We pre-train the network for $800$ epochs with a batch size of $256$. The learning rate is linearly increased to $1e^{-3}$ from 0 in the first $20$ warmup epochs and then decreased to $1e^{-4}$ by the cosine decay schedule. We set the temperature hyperparameter $\tau$ to $0.5$. The experiments are conducted using the PyTorch framework on three NVIDIA TITAN RTX GPUs.

\noindent \textbf{Fine-tuning details.} After the pre-training, we perform fine-tuning. An MLP head is attached to the pre-trained backbone and the whole network is fully fine-tuned for $100$ epochs with cross-entropy loss and a batch size of $64$. The learning rate is linearly increased to $5e-4$ from 0 in the first 5 warm-up epochs and then decreased to $5e-5$ by the cosine decay schedule. To mitigate class imbalance during fine-tuning, we employ a resampling strategy. All hyperparameters are determined through standard coarse-to-fine grid search or step-by-step tuning.

\subsection{Evaluation metrics}
To quantitatively analyze the performance of our method, we use three widely used benchmark metrics for surgical phase recognition: precision, recall, and Jaccard index. Due to phase class imbalance inherent within the EgoSurgery-Phase dataset, the performance will be reported in macro-average. Macro-average is used in imbalanced multi-class settings as it provides equal emphasis on minority classes.

\begin{table}[tb]
\caption{Performance comparison with baseline and state-of-the-art phase recognition models on EgoSurgery-Phase.}
\begin{center}
\resizebox{0.75\textwidth}{!}{
\begin{tabular}{ccccc}
\toprule
Methods & Backbone  & Precision & Recall & Jaccard \\
\hline
PhaseLSTM~\cite{Twinanda2016arXiv} & AlexNet  & 36.3 & 33.1 & 21.9 \\
PhaseNet~\cite{TwinandaTMI2017} & AlexNet   & 37.0 & 25.7 & 19.7 \\
TeCNO~\cite{Tobias2020MICCAI} & ResNet-50    & 47.7 & 39.2 & 27.3   \\
Trans-SVNet~\cite{Gao2021MICCAI} & ResNet-50    & 41.8 & 35.9 & 23.1   \\
NETE~\cite{Yi2023ACCV} & Inception v3    & 43.7 & 35.2 & 27.5  \\
\rowcolor{Gray} GGMAE (Ours) & ViT-S   &  \textbf{51.7}  &  \textbf{45.6} &  \textbf{33.9}  \\
\bottomrule
\end{tabular}}
\end{center}
\label{tab:comparison-with-state-of-the-art}
\end{table}

\subsection{Phase recognition performance comparison}
\noindent\textbf{Comparison with phase recognition methods:}
We first compare our approach with current state-of-the-art phase recognition methods, including TeCNO~\cite{Tobias2020MICCAI}, Trans-SVNet~\cite{Gao2021MICCAI}, and NETE~\cite{Yi2023ACCV}, alongside common baselines PhaseLSTM~\cite{Twinanda2016arXiv} and PhaseNet~\cite{TwinandaTMI2017}. The performance of all methods is summarized in Table \ref{tab:comparison-with-state-of-the-art}. Our GGMAE notably surpasses the baselines in all metrics. Specifically, our method exhibits a substantial improvement over NETE, which is the best performance among previous state-of-the-art methods, by $8.0\%$ (from $43.7\%$ to $51.7\%$) in the Precision, $10.4\%$ (from $35.2\%$ to $45.6\%$) in the Recall, and $6.4\%$ (from $27.5\%$ to $33.9\%$) in the Jaccard index.

\begin{table}[tb]
\caption{Performance comparison with state-of-the-art masked autoencoder-based models on Egosurgery-Phase. The supervised baseline is ViT-S trained from scratch on Egosurgery-Phase.}
\begin{center}
\resizebox{\textwidth}{!}{
\begin{tabular}{cccccc}
\toprule
Methods & Backbone & Masking  & Precision & Recall & Jaccard \\
\hline
Supervised & ViT-S &  & 47.9 & 31.6 & 27.1     \\
VideoMAE~\cite{Tong2022NeurIPS} & ViT-S & Tube masking   & 49.3 & 41.6 & 29.8  \\
VideoMAE V2~\cite{Wang2023CVPR} & ViT-S & Dual masking & \textbf{54.2} & 43.2 & 30.8 \\
SurgMAE~\cite{Muhammad2023arXiv} & ViT-S & Spatio-temporal masking &52.2 & 41.9 & 27.8  \\
\rowcolor{Gray} GGMAE (Ours) & ViT-S & Gaze-guided masking    &  51.7  &  \textbf{45.6} &  \textbf{33.9}  \\
\bottomrule
\end{tabular}}
\end{center}
\label{tab:comparison-with-state-of-the-art-self-supervised}
\end{table}

\noindent \textbf{Comparison with masked autoencoder-based methods.} After being pre-trained with the proposed GGMAE framework, the model exhibits significant performance improvements compared to the model trained from scratch ($6\%$ improvement in the Jaccard index). We then compare current state-of-the-art MAE-based methods, namely VideoMAE and VideoMAEV2. Additionally, we evaluate our approach against SurgMAE, which first demonstrates the effectiveness of MAEs in the surgical domain. The performance of all methods is summarized in Table \ref{tab:comparison-with-state-of-the-art-self-supervised}. Employing the same backbone and training schema, GGMAE surpasses VideoMAE by $4.1\%$ and VideoMAEV2 by $3.1\%$ and  SurgMAE by $6.1\%$ in terms of Jaccard index. 

\subsection{Ablation study}
\noindent \textbf{Mask sampling strategy.} To verify the effectiveness of the proposed gaze-guided masking strategy, we compare its performance with that of random and tube masking. As we can see, our gaze-guided masking strategy brings absolute performance improvements of $3.3\%$. This suggests that the gaze information, as an empirical semantic richness prior, can effectively guide the masking process.

\noindent \textbf{Masking Ratio.} As shown in Tab \ref{tab:ablation-study} (b), we experimented with different masking ratios. Results show that either too large or too small masking ratios have a negative impact on performance. We empirically found that a masking ratio of $90\%$ exhibits the best results.

\noindent \textbf{Temerature parameter.}
We experimented with different temperature parameters $\tau$. As the temperature parameter decreases, the region toward which the gaze is directed becomes more likely to be masked.  As shown in Tab \ref{tab:ablation-study} (c), Our GGMAE exhibits the best performance when temperature parameters $\tau$ is $0.5$. Overall, a temperature parameter $\tau$ is set to $0.5$ by default.
\begin{table*}[tb]
 \caption{Ablation studies on Egosurgery-Phase. We use ViT-S as a backbone for all the experiments.}
\begin{center}
\begin{tabular}{c}
\begin{tabular}{ccc}
{\footnotesize (a) Mask sampling strategy.}&
{\footnotesize (b) Masking ratio ($\rho$)}& {\footnotesize (c) Temperature parameter ($\tau$).}
\\
\begin{tabular}{ccc}
\toprule
Strategy & Ratio &  Jaccard \\
 \hline
Random~\cite{He2022CVPR} & 0.75 & 28.9 \\
Random~\cite{He2022CVPR}  & 0.90 &  30.6\\
Tube~\cite{Tong2022NeurIPS}  & 0.90 & 29.8  \\
\rowcolor{Gray} Gaze-guided & 0.90 &  \textbf{33.9} \\
\bottomrule
\end{tabular}
&
\begin{tabular}{ccc}
\toprule
Ratio & Jaccard \\
 \hline
0.95 & 31.2 \\
\rowcolor{Gray} 0.90 & \textbf{33.9} \\
\rowcolor{white} 0.85 & 31.6 \\
0.80  & 31.5 \\
\bottomrule
\end{tabular}
&
\begin{tabular}{cc}
\toprule
$\tau$  & Jaccard \\
 \hline
1.00 & 30.1  \\
0.75 & 30.6 \\
\rowcolor{Gray} 0.50  & \textbf{33.9} \\
\rowcolor{white} 0.25  & 27.2 \\
\bottomrule
\end{tabular}
\end{tabular} 
\end{tabular}
\end{center}
\label{tab:ablation-study}
\end{table*}

\section{Conclusion and Future Work}
In this paper, we construct the first egocentric open surgery video dataset, Egosurgery-Phase, for phase recognition. We also propose a gaze-guided masked autoencoder, GGMAE, to promote better attention to semantically rich spatial regions using gaze information. Furthermore, GGMAE achieves substantial improvements compared to the existing phase recognition methods and masked autoencoder methods. The remaining challenges for this dataset involve improving model performance on the Egosurgery-Phase. By releasing this dataset to the public, we, alongside the wider research community, aspire to address these challenges in the future collaboratively. Moreover, we intend to enrich this dataset by augmenting the video content and incorporating footage captured from various perspectives (\textit{e.g.}, assistant surgeons, anesthesiologists, perfusionists, and nurses) to advance the automated analysis of open surgery videos.

\subsubsection{\ackname} This work was supported by JSPS KAKENHI Grant Number 22H03617. We would like to thank the reviewers for their
valuable comments.

\subsubsection{Disclosure of Interests.} The authors have no competing interests to declare that are relevant to the content of this article. 

\bibliographystyle{splncs04}
\bibliography{ref}

\end{document}